\documentclass[letterpaper, 10 pt, conference]{ieeeconf}  

\IEEEoverridecommandlockouts                              

\usepackage{graphicx}
\usepackage{caption}
\graphicspath{
    {figures/}
}

\usepackage{amsmath,amssymb,enumerate}

\usepackage{amsthm}
\usepackage{setspace}
\usepackage{booktabs}
\usepackage[usenames,dvipsnames,svgnames,table]{xcolor}
\usepackage{graphics} 
\usepackage{mathtools}
\usepackage{algorithm, algorithmicx, algpseudocode}
\usepackage{blindtext}
\usepackage{gensymb}
\usepackage{xparse}
\usepackage{lipsum}
\usepackage{mathrsfs}
\usepackage[mathscr]{euscript}
\usepackage{times}
\usepackage{cite} 
\usepackage{multicol}
\definecolor{darkgreen}{rgb}{0,0.6,0}
\usepackage[bookmarks=true,colorlinks=true,pdfpagemode=UseNone,citecolor=darkgreen,linkcolor=black,urlcolor=BrickRed,pagebackref]{hyperref}
\usepackage[caption=false,font=footnotesize]{subfig}
\usepackage{amsfonts}
\usepackage[utf8]{inputenc}
\usepackage[T1]{fontenc}
\usepackage{textcomp}
\usepackage{arydshln}
\usepackage{balance}

\renewcommand{\thefigure}{\arabic{figure}}

\definecolor{note}{rgb}{0.1,0.1,1}
\definecolor{rephase}{rgb}{0.15,0.7,0.15}
\definecolor{bag}{rgb}{0.6,0.6,0.2}

\makeatletter
\renewcommand*\env@matrix[1][c]{\hskip -\arraycolsep
  \let\@ifnextchar\new@ifnextchar
  \array{*\c@MaxMatrixCols #1}}
\makeatother

\makeatletter
\newcommand{\mathleft}{\@fleqntrue\@mathmargin0pt}
\newcommand{\mathcenter}{\@fleqnfalse}
\makeatother

\title{\LARGE \bf Deep Convolutional Neural Network and Transfer Learning for Locomotion Intent Prediction} 

\author{Duong Le, Shihao Cheng, Robert D. Gregg, and Maani Ghaffari%
\thanks{Funding for D. Le was provided by Vingroup Scholarship Program. Funding for S. Cheng and R. Gregg was provided by the National Institute of Child Health \& Human Development of the NIH under Award Number R01HD094772. Funding for M. Ghaffari was provided by NSF Award No. 2118818. The content is solely the responsibility of the authors and does not necessarily represent the official views of the NIH or NSF. }
\thanks{The authors are with the College of Engineering, University of Michigan, Ann Arbor, MI 48109, USA. {\tt\small\{duongqle,chengsh,rdgregg,maanigj\} @umich.edu}}%
}

\begin{document}

\maketitle
\thispagestyle{empty}
\pagestyle{empty}

\setlength{\belowdisplayskip}{2pt}
\setlength{\textfloatsep}{4pt}	

\begin{abstract}
Powered prosthetic legs must anticipate the user's intent when switching between different locomotion modes (e.g., level walking, stair ascent/descent, ramp ascent/descent). Numerous data-driven classification techniques have demonstrated promising results for predicting user intent, but the performance of these intent prediction models on novel subjects remains undesirable. In other domains (e.g., image classification), transfer learning has improved classification accuracy by using previously learned features from a large dataset (i.e., pre-trained models) and then transferring this learned model to a new task where a smaller dataset is available. In this paper, we develop a deep convolutional neural network with intra-subject (subject-dependent) and inter-subject (subject-independent) validations based on a human locomotion dataset. We then apply transfer learning for the subject-independent model using a small portion (10\%) of the data from the left-out subject. We compare the performance of these three models. Our results indicate that the transfer learning (TL) model outperforms the subject-independent (IND) model and is comparable to the subject-dependent (DEP) model (DEP Error: 0.74 $\pm$ 0.002\%, IND Error: 11.59 $\pm$ 0.076\%, TL Error: 3.57 $\pm$ 0.02\% with 10\% data). Moreover, as expected, transfer learning accuracy increases with the availability of more data from the left-out subject. We also evaluate the performance of the intent prediction system in various sensor configurations that may be available in a prosthetic leg application. Our results suggest that a thigh IMU on the the prosthesis is sufficient to predict locomotion intent in practice.

\end{abstract} 

\IEEEpeerreviewmaketitle

\section{Introduction}


Emerging powered prosthetic legs have the potential to restore normative gait biomechanics in individuals with limb loss. The most prevalent control framework for powered prosthetic devices has three levels~\cite{Tucker2015, Varol2009}. The high-level controller is responsible for recognizing the user's intent (e.g., level walking, sitting, stair ascent, stair descent) and estimating the environment's parameters (e.g., ground slope, stair height) to provide input to the mid-level controller. The mid-level controller calculates the reference prosthetic joint angles or torques with information from the high-level controller. Finally, the low-level controller tracks the reference signal and drives the motors so that the prosthesis operates according to the user's intentions and task conditions.

\begin{figure}[t]
    \centering
    \includegraphics[width=.99\columnwidth]{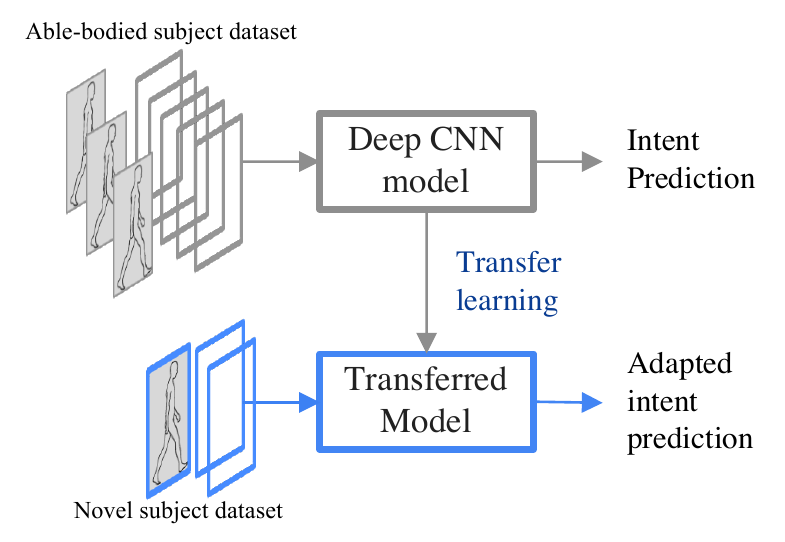}
    \caption{An intent prediction model based on a convolutional neural network trained on a large dataset will not perform well when validating new subject data. We study a novel application of transfer learning to improve the performance of the intent prediction model with limited data from a new subject.}
    \label{fig:my_label}
\end{figure}

It is evident that accurately and effectively predicting user intent is crucial for the prosthetic controller to perform well without negatively impacting the user. For instance, an incorrect classification between specific activities, such as level walking and stair descent, could result in the user falling down the stairs. Recent techniques predict user intent continuously by combining supervised learning algorithms with real-time data from onboard mechanical sensors such as an Inertial Measurement Unit (IMU)~\cite{Bhakta2020, Cheng2021, Bruinsma2021, Su2019}. These investigations highlight the potential of predicting user intent by using modern supervised learning algorithms with onboard sensor data. These algorithms can be divided into two categories: feature engineering and feature learning. For the feature engineering method, manually computed statistical values from sensor data are fed into supervised learning models such as Linear Discriminant Analysis (LD) or Support Vector Machines (SVM). On the contrary, the feature learning method will automatically extract features from sensor data using deep neural networks such as convolutional neural network (CNN), recurrent neural network (RNN), or long short-term memory (LSTM)~\cite{Kuangen2019, Jindong2017}.

Compared to studies using data collected from able-bodied individuals, fewer studies on locomotion intent prediction use data collected from individuals with a limb amputation~\cite{Bruinsma2021}. Some studies use data from amputee participants, but the number of subjects is small, making it difficult to generalize and extend intent prediction models to other users~\cite{Bhakta2020}. Different studies use data from able-bodied datasets, which have more subjects~\cite{Kang2022, Lee2020}. However, when evaluating intent prediction models with subjects not included in the trained dataset, the accuracy is significantly lower than when the trained dataset contains test subject data. In clinical environments, it is challenging to collect sufficient data from amputee patients to train a supervised learning model; therefore, a general intent prediction model that can be applied to the novel subject is desired.

This paper proposes a method to improve the accuracy of a base CNN model evaluated on a novel subject by using transfer learning, which requires only a few strides of training data per activity from that new user. The transfer learning method is driven by the fact that humans can learn more quickly, efficiently, and effectively by transferring knowledge from earlier learning experiences. Transfer learning can reduce the quantity of data needed to train a deep CNN model by using techniques such as feature extraction or modifying parameters from a pre-trained model~\cite{Abolfazl2021, Kaur2021, Pan2010}. 

We use a public able-bodied dataset~\cite{Hu2018} to train and evaluate the intent prediction model based on a deep CNN. The subject-dependent and subject-independent models are developed and trained using data from various sensor options that can be obtained from a prosthetic leg and amputee user. The subject-independent model is transferred to a new model (referred to as the transfer learning model) by freezing the convolutional layers and training the fully connected layers with a small sample of test subject data. This study demonstrates that the error rate of the subject-dependent model is $1.20\pm 0.002$\%, the subject-independent model is $12.87\pm 0.077$\%, and the transfer learning model using 10\% test subject data is $5.28\pm 0.029$\%. The transfer learning model's accuracy increases proportional to the quantity of data included in the training. 

In addition, we demonstrate that the accuracy of using data from a single IMU on the thigh of one leg (e.g., the prosthesis) is comparable to using other sensor configurations. This is significant for practical applications because we do not need to add sensors to the sound leg nor rely on prosthetic leg sensors that are directly affected by the controller rather than the user (i.e., encoders, shank IMU).

The main contributions of this paper are:
\begin{enumerate}
    \item Application of transfer learning to a subject-independent model to increase the accuracy of locomotion intent prediction for new subjects with just a few test subject strides per activity.
    \item Analysis of various sensor configurations suggesting that a thigh IMU on one leg (e.g., the prosthesis) can be used as the input signal to predict locomotion intent with comparable accuracy to other options.
\end{enumerate}


The remaining sections are organized as follows. Section \ref{Method} describes the dataset and intent prediction models. The results are summarized in Section \ref{Results} and discussed in Section \ref{Disscussions}. In Section \ref{Conclusions}, concluding remarks are drawn.


\section{Methods} \label{Method}
This section describes the deep neural network architectures for locomotion intent prediction based on raw data obtained from able-bodied individuals using a variety of sensor options.

\subsection{Dataset}
In this study, we use a publicly available dataset titled ENcyclopedia  of  Able-bodied  Bilateral  Lower  Limb  Locomotor  Signals  (ENABL3S)~\cite{Hu2018} 
to train a deep convolutional neural network model for predicting user intentions. The dataset includes various types of sensor signals mounted on ten different able-bodied subjects, such as wearable electrogoniometers (GONIO), surface electromyography (EMG) and inertial measurement unit (IMU) sensors. Since mechanical sensors are commonly available on powered prosthetic legs \cite{Torrealba2019}, our study focuses on using features from IMU and GONIO to predict the intent of users. 

The GONIO sensor measures the knee and ankle angle/velocity, which would be measured by joint encoders in a powered prosthetic leg. In the dataset, each subject repeats approximately 25 trials of a two-stage circuit comprising of sitting (S), standing (St), level walking (LW), stair ascent (SA) (stage 1) or ramp ascent (RA) (stage 2), LW, ramp descent (RD) (stage 1) or stair descent (SD) (stage 2), LW, and S. The slope of the ramp is 10 degrees, and the staircase has four steps. This dataset labels the locomotion modes manually with a key fob. 
We exclude one of the subjects (AB186) from the dataset because we conjecture there is a problem with the thigh IMU signals, e.g., the sign of the y-axis thigh angular velocity is opposite to that of the other subjects performing the same mode.

\subsection{Data Preparation}
\subsubsection{Data input}


We analyze and evaluate the selection of features gathered from various sensors to predict the user's intentions. Four scenarios are compared, including (i) a thigh IMU on one leg (e.g., the prosthesis), (ii) thigh IMUs on both legs, (iii) onboard sensors available to a prosthesis (i.e., IMUs on the thigh and shank, encoders (ENCs) on the ankle and knee joints), and (iv) all sensors (i.e., including sensors on the prosthesis and an IMU on the healthy leg). The ENC signal correlates with the GONIO sensor signal in the dataset. Each IMU has a 3-axis gyroscope and a 3-axis accelerometer, providing three signals for angular velocity and three for linear acceleration, respectively. 

Previous studies relied on a prior gait event signal window as a data sample to be input into the model~\cite{Lee2020, Bhakta2020}. However, this strategy can be difficult to implement in practice due to the difficulty of identifying gait events~\cite{Xu2021}, and missed gait events could cause missed mode transitions. In this study, a 500 ms sliding window similar to~\cite{Bruinsma2021, Kang2022} is sampled for model input, and the label is the locomotion mode in the next time step.

\subsubsection{Data splitting}
We evaluate three models for predicting user intent: subject-dependent, subject-independent, and transfer learning. 

a) Subject dependent: We train and test the deep learning models using the data from the same subject. For each subject, we split that subject's data into training (80\%), validation (10\%), and testing (10\%).

b) Subject independent: We use leave-one-subject-out cross-validation to train the deep learning models on the data from a set of subjects and test the model on a new untrained subject. The training and validation set consists of data from eight subjects divided into 80\% for training and 20\% for validation. The remaining subject data is used as a test set. This process was repeated nine times until all subjects have been tested.

c) Transfer learning: The transfer learning method is applied to a subject-independent model that uses a portion of the test subject's data for the training process. In order to validate the effect of data size, we use 5\% (3\% for training and 2\% for validation), 10\% (7\% for training and 3\% for validation), 15\% (10\% for training and 5\% for validation), and 20\% (15\% for training and 5\% for validation) from the test subject data for model training. We also apply this process to all subjects with the corresponding subject-independent model.

\subsubsection{Data output}
In our previous studies~\cite{Best2021, cheng2022}, we merge standing and ramp walking as level walking with variable velocities and slopes, thus we simply need to predict sit (S), level walking (LW), stair ascent (SA), and stair descent (SD), as well as the transitions between them: S2W, W2S, W2SA, SA2W, SD2W. In total we need to classify ten different locomotion modes.

\subsection{Deep Neural Networks Model Architectures}
\subsubsection{Base convolutional neural network}

\begin{figure} 
  \centering
  \includegraphics[width=0.8\columnwidth]{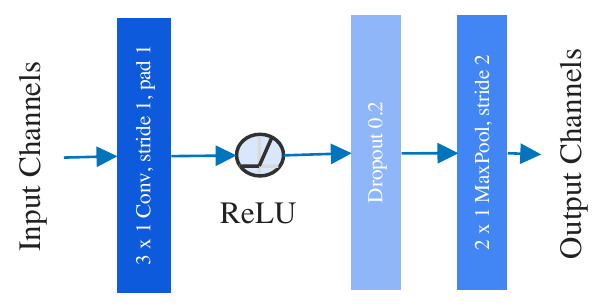}  
  \caption{\label{fig:vgg_block} VGG block architectures consist of one convolutional (Conv) layer with 3×1 kernels with padding of 1 and stride of 1 (keeping the height and width) followed by ReLU activation function, dropout of 0.2, and a 2×1 max-pooling layer with a stride of 2 (halving height after each block). The convolutional part of the network connects several VGG blocks. 
  The input channels to the first block correspond to sensor signals. For example, each IMU sensor has six channels and each encoder has two channels.}
\end{figure}

We build the intent prediction models with a deep convolutional neural network. We train and validate the CNN models with two different methods, i.e., subject-dependent and subject-independent. CNN model is developed with the VGG network method~\cite{Simonyan2014, zhang2021dive}. The model is built from VGG blocks, with each VGG block beginning with a convolutional layer and ending with a max pooling layer as shown in Fig.~\ref{fig:vgg_block}. The convolutional layer is designed to extract complex features from the input data, and the one-dimensional max pooling layer is added after each block to reduce the data sample size. A one-dimensional kernel is used to enhance computational performance in terms of memory usage and execution time. The rectified linear activation function (ReLU) is used for the convolutional layer, together with a dropout mechanism to prevent overfitting.

\begin{figure} 
  \centering
  \includegraphics[width=0.9\columnwidth]{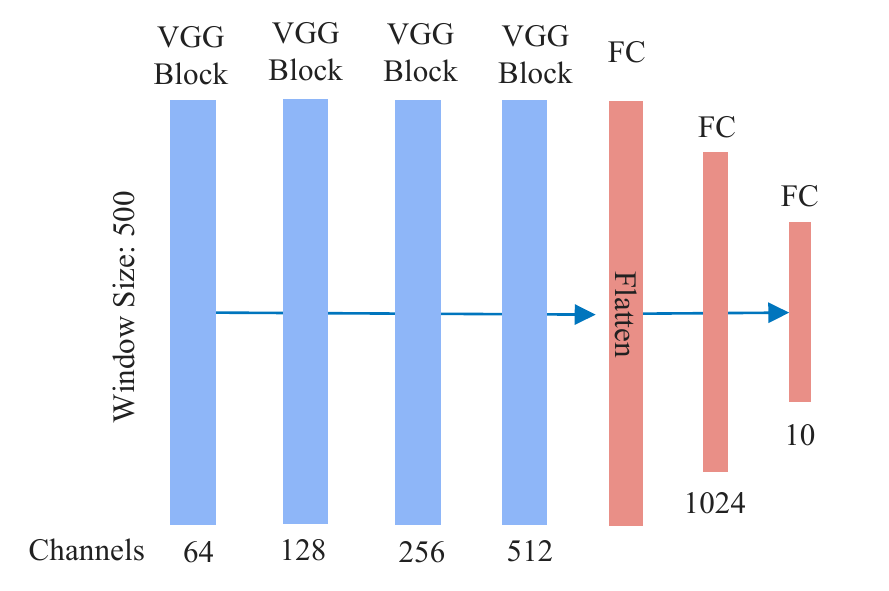}  
  \caption{\label{fig:dep_cnn} Subject-dependent model architectures include four continuously connected VGG blocks and three fully connected layers. The output channels for each VGG block are 64, 128, 246, and 512. The first fully connected layer flatten the output from convolution layers. The output size of second fully connected layer is 1024 and the last layer is ten, which is the number of labeled modes.}
\end{figure}

\begin{figure} 
  \centering
  \includegraphics[width=0.9\columnwidth]{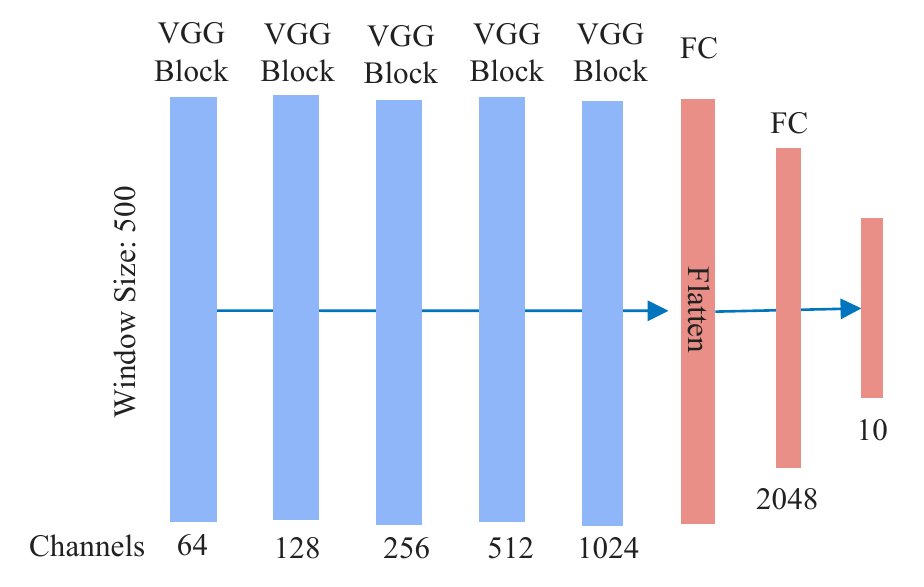}  
  \caption{\label{fig:ind_cnn} Subject-independent model architectures included five continuously connected VGG blocks and three fully connected layers. The output channels for each VGG block are 64, 128, 246, 512, and 1024. The first fully connected layer flatten the output from convolution layers. The output size of second fully connected layer is 1024 and the last layer is ten, which is number of labeled modes.}
\end{figure}

As depicted in Fig. \ref{fig:dep_cnn}, the model employed for the subject-dependent method consists of four continuously connected VGG blocks. The number of input channels in the first block depends on the number of sensors used; for instance, an IMU sensor has six channels, whereas an encoder has two. Each VGG block's output channels correspond to the input of the following block. A series of 64–128–256–512 output channels are developed for the subject-dependent model. As shown in Figure \ref{fig:ind_cnn}, the subject-independent model uses one more VGG block than the subject-dependent model, with output channels of 1028. The VGG blocks are interconnected to construct a system of convolutional layers responsible for extracting data-descriptive features.

Features learned by the convolutional layers are then fed forward into the fully connected (FC) layers, which are used to map those features into the pre-labeled ten locomotion modes. Similar to convolutional layers, we also involve dropout methods and the ReLU activation function in the first two fully connected layers to prevent overfitting. The second fully connected layer in the subject-dependent model has 1024 outputs while subject-independent has 2048 outputs. Both models have the fully connected output layer with 10 outputs, which is the number of labeled locomotion modes. 

\subsubsection{Transfer learning}

The subject-independent model is then refined via a transfer learning approach by freezing the learned parameters in the convolutional layers and relearning the parameters in the fully connected layers.

\subsection{Deep Neural Networks Model Training}
This section covers how to train deep convolutional neural networks. The training was conducted on a PC with 2 NVIDIA TITAN RTX GPUs. The model was constructed and trained using the Pytorch Lightning Library.

\subsubsection{Loss function}

We set up a classification problem by minimizing the cross entropy loss defined as $$L=-\sum_{c=1}^C w_c \log \frac{\exp \left(x_c\right)}{\sum_{i=1}^C \exp \left(x_i\right)} y_c,$$ where \(y_c\) is the target, \(x_c\) is the input, \(w_c\) is the class weight, and \(C\) is the number of classes.



The number of data samples from different classes (i.e., locomotion modes) is unbalanced in the ENABL3S dataset. To address this issue, we assign different weights to each class based on the percentage of strides for each activity and account that into the loss function.

\subsubsection{Hyperparameter}

We investigate various CNN model hyperparameters for optimal performance. The number of epochs, batch size, and learning rate are involved parameters in model training. Table \ref{training_params} contains the parameters chosen for high accuracy.

The training data are divided into data sample batches for model training. The batches are used in succession until all training data is consumed to conclude an epoch. The process is repeated numerous times, and the greater the number of epochs, the higher the training accuracy but also the longer the training time.

Learning rate is another parameter for model training. The higher the learning rate, the quicker the training, but if it is too high, the model cannot converge. Conversely, if the learning rate is too low, the loss function may decrease to a local minimum, or the training time may become too long.

\begin{table}[t]
\caption{Optimal hyperparameter (i.e., number of epoch, batch size, and learning rate) for different CNN models included subject-dependent (Dep.), subject-independent (i.e., Ind.), and transfer learning (i.e., Transfer).}
\label{training_params}
\footnotesize
\begin{center}
\begin{tabular}{c|c|c|c}
\hline
& Dep. & Ind. & Transfer \\
\hline
Epoch & 30& 35& 100\\
\hline
Batch size& 512& 1024& 256\\
\hline
Learning rate & 0.0001& 0.00015& 0.0001\\
\hline
\end{tabular}
\end{center}
\end{table}

\subsubsection{Optimizer}
Depending on the optimizer's formula, the optimizer can be interpreted as a mathematical function that adjusts the neural network's weights using gradients and additional information. We use the Adaptive Moment Estimation (Adam)~\cite{Kingma2014} optimizer for the subject-dependent and subject-independent models and the stochastic gradient descent (SGD) algorithm for the transfer learning model in this study. Adam is the default optimization algorithm for training deep learning models in numerous studies due to its faster training speed. However, when the transfer learning method is used, the amount of training data and number of parameters are small, so the more efficient SGD method is the most suitable option~\cite{Sun2019, Sun2019_2}.

\subsection{Evaluation}

Similar to previous studies on locomotion intent classifiers~\cite{Hu2018, Bhakta2020, Cheng2021, Lee2020}, we evaluate the model based on the error rate criterion determined by the proportion of incorrect predictions relative to the total number of test data samples. Moreover, the error rates can be categorized as steady state (SS) and transition (TS). A TS occurs when the current mode differs from the previous mode, whereas a SS occurs when the current mode is identical to the previous mode. The formula for calculating these criteria are:

\begin{equation}
\text{Overall\_error}=1 - \frac{\text{Overall\_correct}}{\text{Overall\_total}},
\end{equation}

\begin{equation}
\text{SS\_error}=1 - \frac{\text{SS\_correct}}{\text{SS\_total}},
\end{equation}

\begin{equation}
\text{TS\_error}=1 - \frac{\text{TS\_correct}}{\text{TS\_total}}.
\end{equation}

In addition, we perform statistical analysis to compare the model performance over two factors: learning condition (DEP, IND, and Transfer Learning) and sensor setup (unilateral thigh, bilateral thigh, prosthetic sensors, and all). To that end, we conduct a two-way repeated measures analysis of variance (ANOVA) with classification error as the dependent variable, and learning condition and sensor setup as independent variables. Then, we followed up with a Bonferroni post-hoc analysis for the significant factors to determine the statistical significant differences (p < 0.05) between each level within each factor. Finally, we ran multiple pairwise comparisons to calculate the significant difference in error rate between each specific pair of learning conditions within each sensor setup (e.g., between unilateral thigh under DEP and unilateral thigh under IND).

\section{Results} \label{Results}
\begin{figure}[t]
  \centering
  \vspace{-80pt} 
  \includegraphics[width=0.48\textwidth]{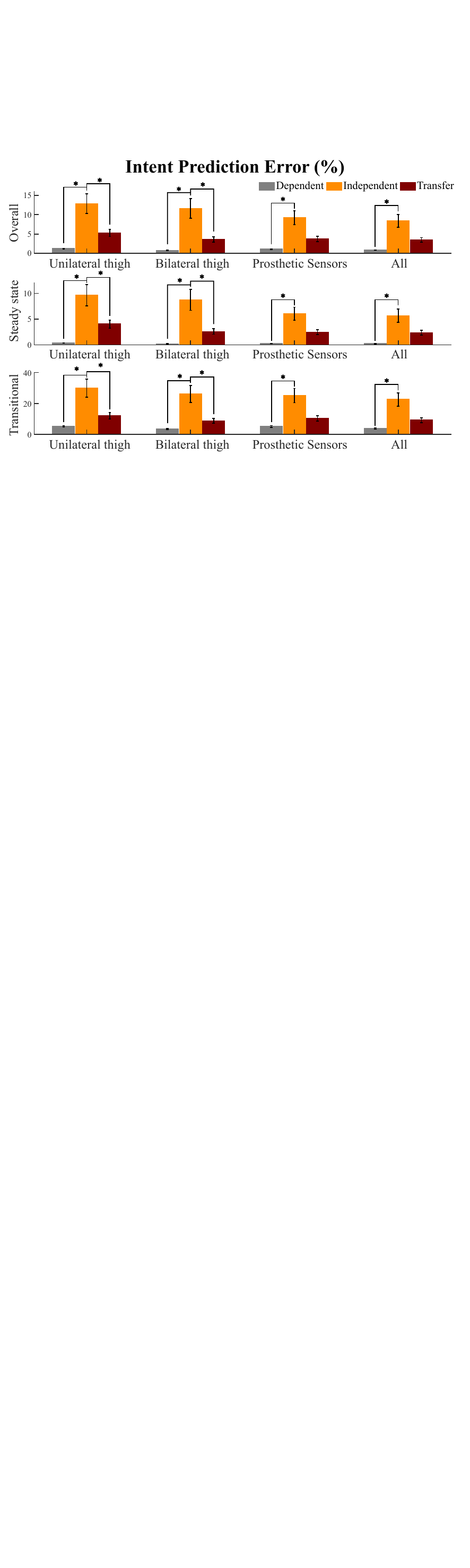}
  \vspace{-585pt} 
  \caption{Three different models (subject-dependent, subject-independent, and transfer learning) are compared in different sensor setups and error types. The unilateral thigh means using one thigh IMU on the prosthesis leg, the bilateral thigh means using both thigh IMU on both legs, prosthetic sensors means using all sensor on the prostheses, and All means using all sensors included thigh IMU on healthy leg and sensors on prosthetic leg. Error types are calculated as in equations (1)-(3). Error bars represent $\pm$ standard error of the mean. Asterisks indicate statistical significance (p < 0.05).}
  \label{fig:mainResults}
\end{figure}

\begin{figure} 
  \centering
  \includegraphics[width=1\columnwidth]{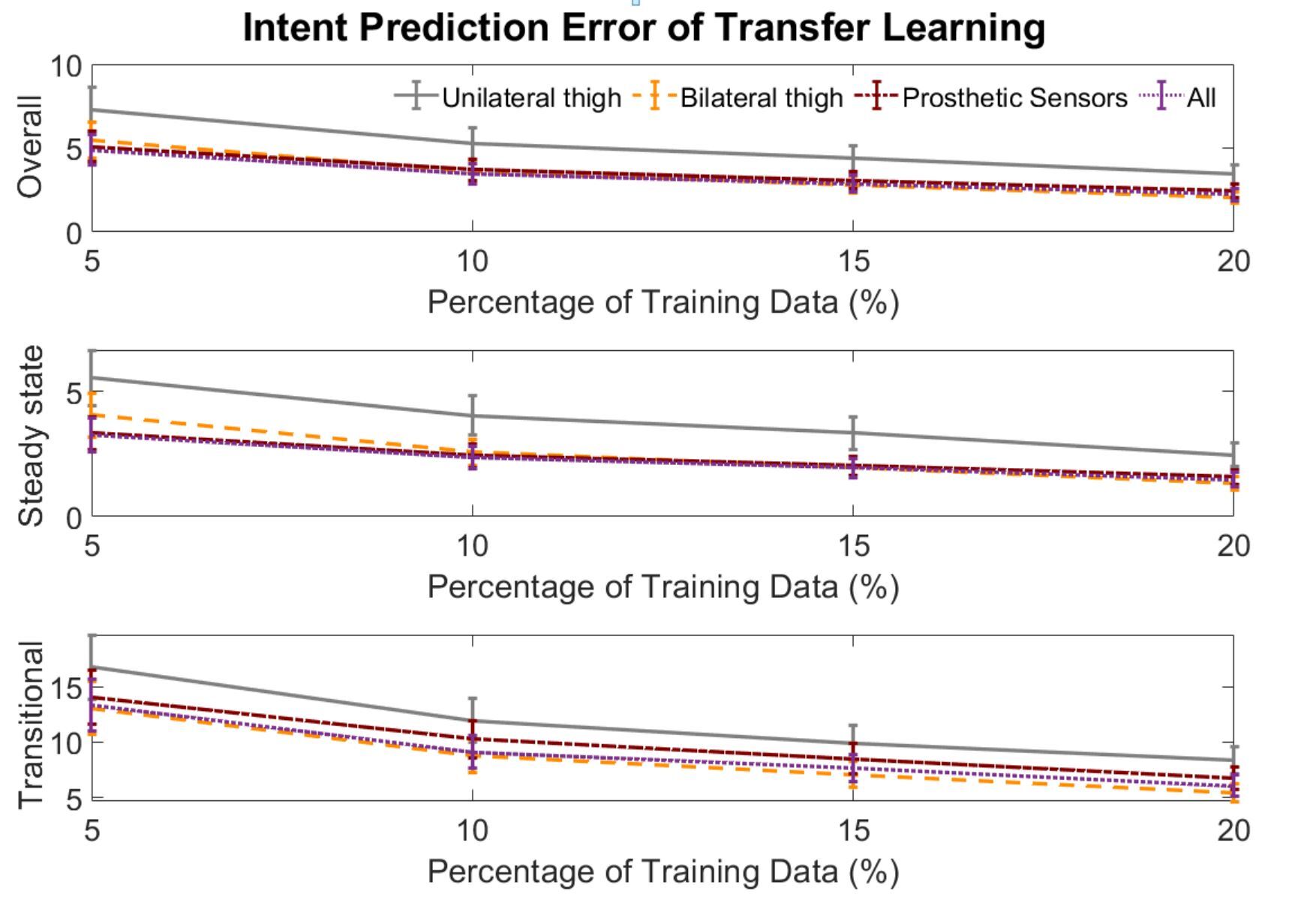}  
  \caption{\label{fig:TransferResults} Prediction error for transfer learning at different percentages of trained dataset from test subject.}
\end{figure}

The results of the three proposed models with different sensor configurations are shown in Fig. \ref{fig:mainResults}, where 10\% of the test subject data is used for the transfer learning method. Moreover, Fig. \ref{fig:TransferResults} shows the effect of the amount of test subject data used in the transfer learning method (5\%, 10\%, 15\%, and 20\%). 

\subsection{Model Comparison}
Table \ref{tbl:result_num} compares the error rates of the intent prediction models with different sensor selections. The overall error rate of the subject-independent model is statistically higher than the subject-dependent model ($p<0.05$), however, the subject-dependent model is not significantly different from the transfer learning model ($p>0.05$). This trend tends to hold with steady state and transition errors for all sensor setups. In addition, transfer learning model demonstrates significantly lower error rate than the subject-independent model only in unilateral thigh and bilateral thigh sensor setups, i.e., no significant differences in Prosthetic and All sensors setups.

\begin{table}[t]
\caption{Error rates of the classifiers on different sensor setups and models including subject-independent (Ind.), subject-dependent (Dep.), and transfer learning (Transfer). All results are represented as mean [standard deviation].} 
\label{tbl:result_num}
\footnotesize
\begin{center}
\begin{tabular}{c|c|c|c|c}
\hline
Sensor setup & Error & Dep. & Ind. & Transfer. \\
\hline
Unilateral Thigh & Overall & 1.20[.002] & 12.87[.077] & 5.28[.029] \\
 & SS & 0.38[.001] & 9.64[.062] & 4.03[.023] \\
 & TS & 5.01[.011] & 29.82[.175] & 11.94[.060] \\
 \hline
Bilateral Thigh & Overall & 0.74[.002] & 11.59[.076] & 3.57[.020] \\
 & SS & 0.18[.001] & 8.72[.061] & 2.57[.016] \\
 & TS & 3.46[.010] & 26.06[.166] & 8.74[.045] \\
 \hline
Prostheses Leg & Overall & 1.06[.003] & 9.19[.053] & 3.71[.019] \\
 & SS & 0.23[.001] & 6.03[.038] & 2.43[.014] \\
 & TS & 5.00[.017] & 25.04[.138] & 10.24[.045] \\
 \hline
All & Overall & 0.84[.002] & 8.41[.049] & 3.45[.018] \\
 & SS & 0.21[.001] & 5.65[.038] & 2.36[.014] \\
 & TS & 3.84[.012] & 22.52[.128] & 9.11[.045] \\
 \hline
\end{tabular}

\end{center}
\end{table}

\subsection{Sensor Selections Comparison}

When we compare models with different sensor choices, ANOVA found no statistical difference between them. However, it is still possible to see some differences between these models. For the subject-independent model, the more sensors used, the lower the overall error rate (Unilateral Thigh: 12.87\%, Bilateral Thigh: 11.59\%, Prostheses Leg: 9.19\%, All: 8.41\%). The same goes for steady-state and transitional error rates. However, with the subject-dependent model, choosing the bilateral thigh sensor option gives the lowest error rate (Overall: 0.74\%, Steady-state: 0.18\%, Transition: 3.46\%). For the transfer learning model, choosing to use all sensors gives the lowest overall and steady-state error rates (Overall: 3.45\%, Steady-state: 2.35\%). But the transfer learning model with Bilateral Thigh sensor selection gives the lowest transition error rate (8.74\%).

\subsection{Data for Transfer Learning Comparison}


\begin{table}[t]
\caption{Estimation of number of strides for 10\% of a subject data. These values are approximated by 10\% of the averaged total number of strides for ten subjects reported in~\cite{Hu2018}.}
\label{tbl:10percentsub}
\footnotesize
\begin{center}
\begin{tabular}{c|c|c}
\hline
 &  & Number of strides \\
 \hline
Steady-state & Level walking (LW) & 45 \\
 & Ramp ascent (RA) & 14 \\
 & Ramp descent (RD) & 18 \\
 & Stair ascent (SA) & 5 \\
 & Stair descent (SD) & 5 \\
 \hline
Transition & LW to RA & 2 \\
 & LW to RD & 2 \\
 & LW to SA & 2 \\
 & LW to SD & 2 \\
 & RA to LW & 2 \\
 & RD to LW & 2 \\
 & SA to LW & 2 \\
 & SD to LW & 2 \\
 \hline
Total &  & 106 \\
\hline
\end{tabular}
\end{center}
\end{table}

Fig. \ref{fig:TransferResults} shows the accuracy of user intent prediction with different percentages of data used in transfer learning for the test subject. Generally, the prediction errors decreased with more training data available from the test subject, and this trend is consistent across different sensor setups. 
Since the ENABL3s dataset contains 25 repetitions of a circuit for each subject, 20\% of the test subject data corresponds to approximately five repetitions and 5\% to more than one repetition. More specifically, Table \ref{tbl:10percentsub} shows an estimation of the number of strides for each locomotion mode with 10\% of one subject's data from~\cite{Hu2018}. This means that 10\% of one subject's data is about 106 strides, including two strides for each mode transition.

\section{Discussions} \label{Disscussions}
We propose using transfer learning to enhance a subject-independent model's ability to predict the locomotion intent of a new subject. In addition, we examine various sensor options for predicting user intent to identify a practical solution.
\subsection{Comparison to Prior Studies}
In this study, we explore the novel application of transfer learning to studies of locomotion intent prediction using mechanical sensors.
It is difficult to directly compare our intent prediction models to previous research due to differences in the training dataset, labeled modes, sensor setups, and evaluation methods. However, we can compare our subject-dependent and subject-independent models to other studies to give context for our transfer learning results. Our overall accuracy in predicting locomotion intent using a single thigh IMU (DEP error rate: 1.2\%) is comparable to the study using the WaveNet model with the same sensor input (DEP error rate: 2.12\%)~\cite {Lu2020}. In addition, our subject-dependent error rate using two thigh IMUs (DEP error rate: 0.74\%) outperforms the study using the same dataset (DEP error rate using only IMU sensors: 2.58\%)~\cite{Lee2020}. Our overall error rate with prosthetic sensor setup (DEP: 1.06\%, IND: 9.19\%) are comparable to the study using XBoost for data from amputees with all sensors on the prosthesis (DEP error: 2.93 \%, IND error: 10,12 \%)~\cite{Bhakta2020}. The similarity of the DEP and IND models compared with previous studies demonstrate that our transfer learning results are meaningful.

\subsection{Implementation of Transfer Learning}
This study demonstrates the effectiveness of transfer learning in improving the accuracy of intent prediction for novel subjects. With the abundance of data collected from able-bodied individuals~\cite{Camargo2021, Reznick2021}, it is straightforward to construct a generalized intent prediction model. In studies on image classification, numerous models are built with vast quantities of data and are made available for use in other research. Then, other researchers who wish to apply image classification to a significantly smaller dataset can use transfer learning for the public pre-trained models. We propose a similar approach to building an intent prediction model by making the dataset public so that various deep learning models can be built and compared. The pre-trained models can then be used based on the actual application, such as constructing an intent prediction model for prosthesis controllers.
In a clinical environment, prosthesis controllers can be tuned to individuals in a pre-designed environment with assistance from a prosthetist~\cite{quintero2018intuitive}. The application of transfer learning can enhance the intent prediction model for individuals operating in an unknown environment. This study demonstrates that the prediction model has improved significantly with about 100 strides, of which only about two are for each transition task (the most difficult tasks to perform).

\subsection{Sensor Selection}
Previous research on predicting locomotion intent for prosthetic control has used all the sensors available on the device~\cite{Bhakta2020}. The prediction accuracy in the simulation is high due to the use of numerous features from sensors. However, in practice, sensors attached to the prosthesis (such as encoders, shank IMU, and foot IMU) are directly influenced by the controller, so they may send signals that differ from the user's intent if the controller is malfunctioning. Furthermore, different research groups use different control strategies for the prosthesis to control the knee and ankle joints, resulting in different extracted features from the sensors on the prosthesis. Therefore, it is challenging to use a model trained on data collected by one research group for a controller developed by another and signal sources directly controlled by the amputee's body are best able to inform user intent. For example. the IMU mounted at the connection point between the prosthetic knee and the amputee socket (i.e., the thigh IMU) can be used for this purpose~\cite{Bartlett2018, Chinimilli2019, Cheng2021}.
However, adding additional sensors (e.g., another thigh IMU) to the user's sound leg is a potential barrier to clinical deployment. According to the statistical analysis, there is no significant difference between using one thigh IMU mounted on the prosthesis (unilateral thigh) versus using thigh IMUs on both legs (bilateral thigh), hence, it is sufficient to predict user intent accurately by using the on-board thigh IMU on the prosthesis.

\subsection{Limitations}
Our study is limited by the small number of subjects in the dataset (N=9), resulting in a high error rate for the subject-independent model, which affects the accuracy of the transfer learning model. The accuracy of the models can be enhanced as more comprehensive datasets become available. 
Similarly, this study only uses data from able-bodied subjects, therefore, the ability for the model to adapt to new amputee subjects requires further validation. In the future work, we will use additional data from amputee subjects to increase the accuracy of the subject-independent model, allowing the transfer learning model to predict amputees intentions with greater accuracy. 

\section{Conclusions} \label{Conclusions}
Our study investigated transfer learning to improve the accuracy of a CNN-based model for predicting the locomotion intent of new subjects. We found that the transfer learning method significantly decreased the error rate of the user-independent model with only about 100 strides (with two strides for each transition task) from the test subject. This error rate decreases as more data from the test subject is used to train the transferred model. We also tested the intent prediction model with different sensor configurations. We concluded that using a thigh IMU on a single leg (e.g., the prosthesis) will provide accuracy comparable to other sensor setups.

{\small 
\balance
\bibliographystyle{IEEEtran}
\bibliography{bib/strings-abrv,bib/ieee-abrv,bib/references}
}

\end{document}